# Framing, Judging, Steering: An Assessable Competency Model for Teaching Students to Reason With Generative AI

**Alexander Apartsin[1] and Yehudit Aperstein[2]**

*[1] Holon Institute of Technology · [2] Afeka College of Engineering*

**Abstract**

Generative AI makes answers easy and understanding hard, and uncritical use invites cognitive offloading. Schools still measure *unaided* performance, yet the real task is to produce good work *with* AI: framing an ill-defined task, judging the output, and steering the model toward a better result. This ability is rarely assessed in its own right; where measured, it collapses into one "prompting" score that cannot diagnose *why* AI use succeeds or fails. We propose **CoRe-3** (Co-Reasoning), a competency model factoring productive AI use into three assessable skills we abbreviate **FJS**: **Framing** (specifying an ill-defined task before invoking AI), **Judging** (evaluating output for errors and unstated assumptions), and **Steering** (iteratively redirecting the model). Its distinguishing claim is the separation of pre-generation Framing from post-generation Steering, with Judging as the gate between. We ground the skills in theory, state five testable propositions, and instantiate them in **CoReasoningLab**, an open platform that presents flawed AI output and scores them independently. Over simulated learners (generated and graded by different models), the skills dissociate: each tracks its own manipulated competence while staying flat in the others, and grades become correlated when one competence is shared across all three (convergent and discriminant validity), across grader backends from two providers. Human-rater agreement and outcomes are next; we release the instrument, data, and protocol.

## 1. Introduction

Most AI-in-education tools optimize for the wrong variable. They shorten the path from question to answer, when the answer was never the point of education. The cognitive work of specifying a problem, evaluating a candidate solution, and improving it is where learning happens; when that work is delegated wholesale to a machine, the learner walks away with a correct artifact and an unchanged mind. The educational opportunity of generative AI is therefore not to deliver answers faster, but to make the reasoning around them visible, practiced, and assessable.

Recent evidence makes the stakes concrete. Frequent, uncritical AI use correlates with lower critical-thinking performance, an effect mediated by cognitive offloading and most pronounced in younger users (Gerlich, 2025). Controlled studies of AI-assisted writing report "metacognitive laziness," in which learners bypass the self-regulatory processes of diagnosing, evaluating, and revising (Fan et al., 2025), and neurophysiological work describes an accumulation of "cognitive debt" when an assistant carries the reasoning load (Kosmyna et al., 2025).

At the same time, field studies of professional AI use show that the benefit of AI is sharply conditional on the user's skill in directing it: assistance helps inside a model's competence frontier and *harms* outside it (Dell'Acqua et al., 2023), and the most effective users adopt an iterative, critical "push-back-and-validate" mode rather than wholesale delegation (Randazzo et al., 2025). Meta-analytic evidence underscores the stakes: human-AI teams frequently underperform the better of the human or the AI alone (Vaccaro et al., 2024), which makes the human's skill in directing the system, not mere access to it, the decisive variable. Direct behavioral evidence sharpens the point: across 16,851 real student-ChatGPT interactions from a university course (McNichols et al., 2025), students' moves are dominated by framing-type queries, while explicit verification of the AI's output (about 4% of interactions) and corrective editing (about 2.5%) are rare, the very under-judging and under-steering a competency model must target.

These findings share a structure. The difference between productive and counterproductive AI use is not access to AI; it is a *competency* that some learners exercise and others do not. This echoes the long-standing distinction between the effects obtained *with* a technology during use and the cognitive residue left *of* it, which depends on the learner's mindful engagement in the partnership (Salomon et al., 1991). If that competency can be named and decomposed, it can be taught and assessed. Current AI-literacy assessment can often tell us *that* a learner's AI use is unproductive but not *why*: whether they specified the task badly, failed to detect the flaws in the output, or knew the flaws but could not correct them. These are different failures with different remedies, and a single "prompting" score conflates them.

This trajectory is not transient, and that is what raises the stakes for assessment. As models absorb more of the execution, the human contribution does not disappear; it concentrates in the operations a system cannot perform on its own behalf: choosing which problem is worth solving and specifying it, judging whether a fluent output is actually right, and redirecting the model when it is not. The field evidence that AI's value is conditional on the user's skill (Dell'Acqua et al., 2023; Vaccaro et al., 2024) implies that this conditioning tightens rather than loosens as systems improve, so the durable, assessable human skill shifts from executing a task to framing, judging, and steering it (Acar, 2023; Denny et al., 2024) (Figure 1). An assessment that still measures unaided execution therefore measures a shrinking part of what competent work will require.

**Where competent work concentrates as models improve**

*schematic; the axes are conceptual, not quantitative*

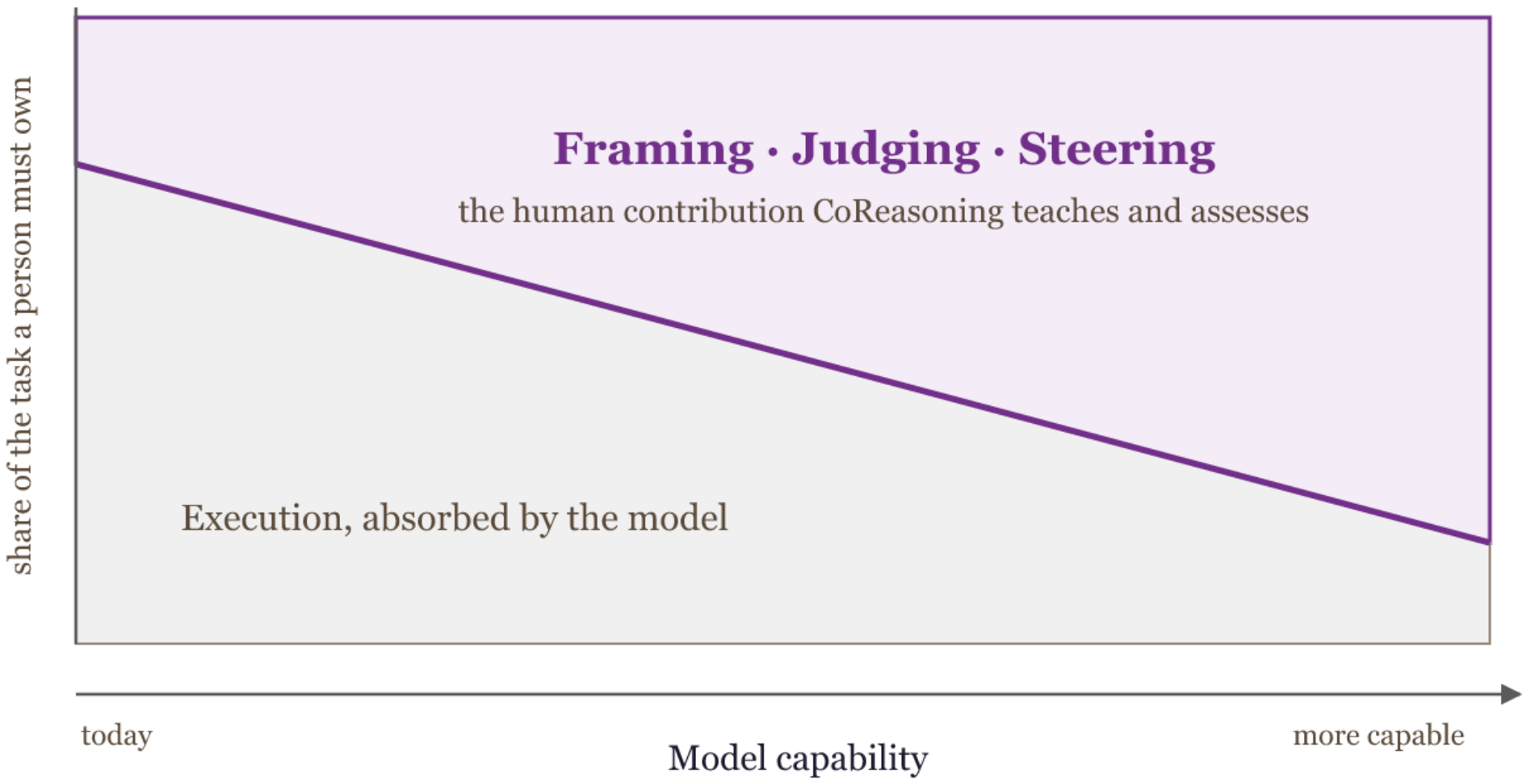

***Figure 1 (schematic).*** *As models absorb more of the execution, the part of a task a competent person must still own shifts from doing the work toward framing, judging, and steering it; the axes are conceptual, not quantitative. CoRe-3 is a decomposition and assessment of that shifting contribution.*

We propose **CoRe-3** (Co-Reasoning), a competency model that decomposes productive work with generative AI into three distinct skills, each independently assessable and collectively abbreviated **FJS**:

- **Framing.** Before invoking the AI, transform an ill-defined problem into a well-specified task: surface unstated assumptions, fix constraints and scope, and define what an adequate solution must satisfy. Framing is a problem-structuring skill exercised *prior to* generation.
- **Judging.** Given AI output, critically evaluate it as a knowledge claim: detect factual and logical errors, missing edge cases, unstated assumptions, and risks, and calibrate how far to rely on it. Judging is an evaluative, epistemic skill.
- **Steering.** Across one or more cycles, issue targeted corrective feedback that moves the AI's output measurably closer to an adequate solution. Steering is a control skill exercised *after* generation, in a loop with Judging.

**Contributions.** This paper makes the following contributions:

1. **A competency model** that decomposes productive work with generative AI into three temporally and cognitively distinct, independently-assessable skills (Framing, Judging, and Steering), whose defining novelty is the separation of the pre-generation skill (Framing) from the post-generation corrective skill (Steering) that prior frameworks fuse.
2. **A theoretical grounding** that maps each skill to established theory (metacognitive monitoring and control, self-regulated learning, epistemic vigilance and critical thinking, productive struggle) under a single unifying monitor-control architecture, so each skill carries an explicit theoretical warrant.
3. **Five testable propositions** about how the skills relate (framing gates the loop; judging gates steering; the skills dissociate; judging transfers less than framing; the model inverts cognitive offloading), proposing new relationships among the constructs (Gilson & Goldberg, 2015; Jaakkola, 2020).
4. **A precise novelty positioning** against the nearest prior frameworks (AI-fluency, metacognitive-demands, prompt-literacy, and AI-literacy models), delineating what the model contributes beyond each.

5. **An instrument and feasibility demonstration** showing that the three skills can be scored automatically, with a multitrait-multimethod validity analysis (three grader backends across two vendors; shared- versus independent-competence populations) giving convergent and discriminant evidence that the three skills are separately measurable and genuinely distinct.
6. **A reproducible, released artifact** (the assessment instrument and the experiment data) together with a fully prepared human-rater validation protocol for the community to run.

The empirical material is a feasibility demonstration of *construct separability and measurability*; the efficacy study that tests learning outcomes is the next stage of the program (Section 9.3).

## 2. The problem with current AI-literacy assessment

Existing instruments for assessing how students work with AI are not wrong so much as under-resolved. Knowledge-oriented frameworks such as Long and Magerko (2020) and the UNESCO (2024) student competency framework define AI literacy chiefly as *understanding AI systems*: what AI is, what it can and cannot do, how it works, and how it should be governed. These are necessary, but they say little about the *performance* of working with a model on a task. Prompt-literacy and generative-AI-literacy models (Lo, 2023; Annapureddy et al., 2024) move closer to performance, but they bundle task specification and iterative refinement into a single "prompting" competency and treat evaluation of the output as a downstream check rather than a co-equal skill.

The cost of this coarse resolution is diagnostic. When a learner's AI use produces a poor result, a single prompting score cannot tell an instructor *which* cognitive operation failed: did the learner specify the task badly, so that the model solved the wrong problem; did they fail to detect the flaws in an otherwise plausible output; or did they see the flaws but issue corrections too vague to fix them? These are three different failures with three different remedies (teaching problem specification, teaching critical evaluation, and teaching corrective communication), and an instrument that cannot separate them cannot guide instruction. Integrative reviews of AI literacy after generative AI confirm that the field still lacks a scheme isolating distinct, independently-assessable reasoning competencies (Gu & Ericson, 2025), even as syntheses document generative AI's mixed effects on critical and creative thinking (Li et al., 2026). A diagnostic competency model must therefore decompose the human-AI loop into the distinct operations that can each break, and must show that those operations are in fact separable in learners. That is the gap CoRe-3 addresses.

## 3. The CoRe-3 framework

### 3.1 A monitor-control architecture with an upstream task-definition

The three FJS skills are not an arbitrary list; they instantiate a well-understood cognitive architecture. Nelson and Narens (1990) describe metacognition as a two-level system: an object level (cognition itself) and a meta level (a dynamic model of the object level), linked by **monitoring** (information flowing from object to meta) and **control** (commands flowing from meta to object). In CoRe-3, the learner's meta level supervises an object level that is *external*, the AI's generative process, rather than the learner's own cognition. This is a deliberate extension of the Nelson-Narens architecture from intrapersonal monitoring to what we term *exo-directed* monitoring and control, and it is the framework's central theoretical move: the same metacognitive machinery is turned outward onto a fallible cognitive artifact. This extension is non-trivial. Exo-directed monitoring adds an epistemic-vigilance burden that self-monitoring does not have: the learner must judge the reliability of a source whose competence differs from, and is hidden from, their own. This is exactly why Judging is bounded by domain knowledge (Proposition P4). With that point made, the mapping is direct:

- **Judging is monitoring.** The learner compares AI output against an internal model of an adequate solution and registers discrepancies.
- **Steering is control.** The learner acts on the monitoring signal, issuing commands that change the object-level process.
- **Framing is task definition.** Before monitoring can occur, the learner must establish the *standards* against which output is judged. In the COPES model of self-regulated learning (conditions, operations, products, evaluations, and standards; Winne & Hadwin, 1998), task definition is the explicit first phase, and products are evaluated against internally held standards; mismatch triggers reprocessing. Framing is that task-definition phase applied to a human-AI loop: it sets the referent that makes Judging and Steering well-posed.

The Judge→Steer cycle is therefore a monitor→control loop seeded by a task definition. This is the structural spine of the framework and the reason the three skills cohere rather than coexist by stipulation (Figure 2).

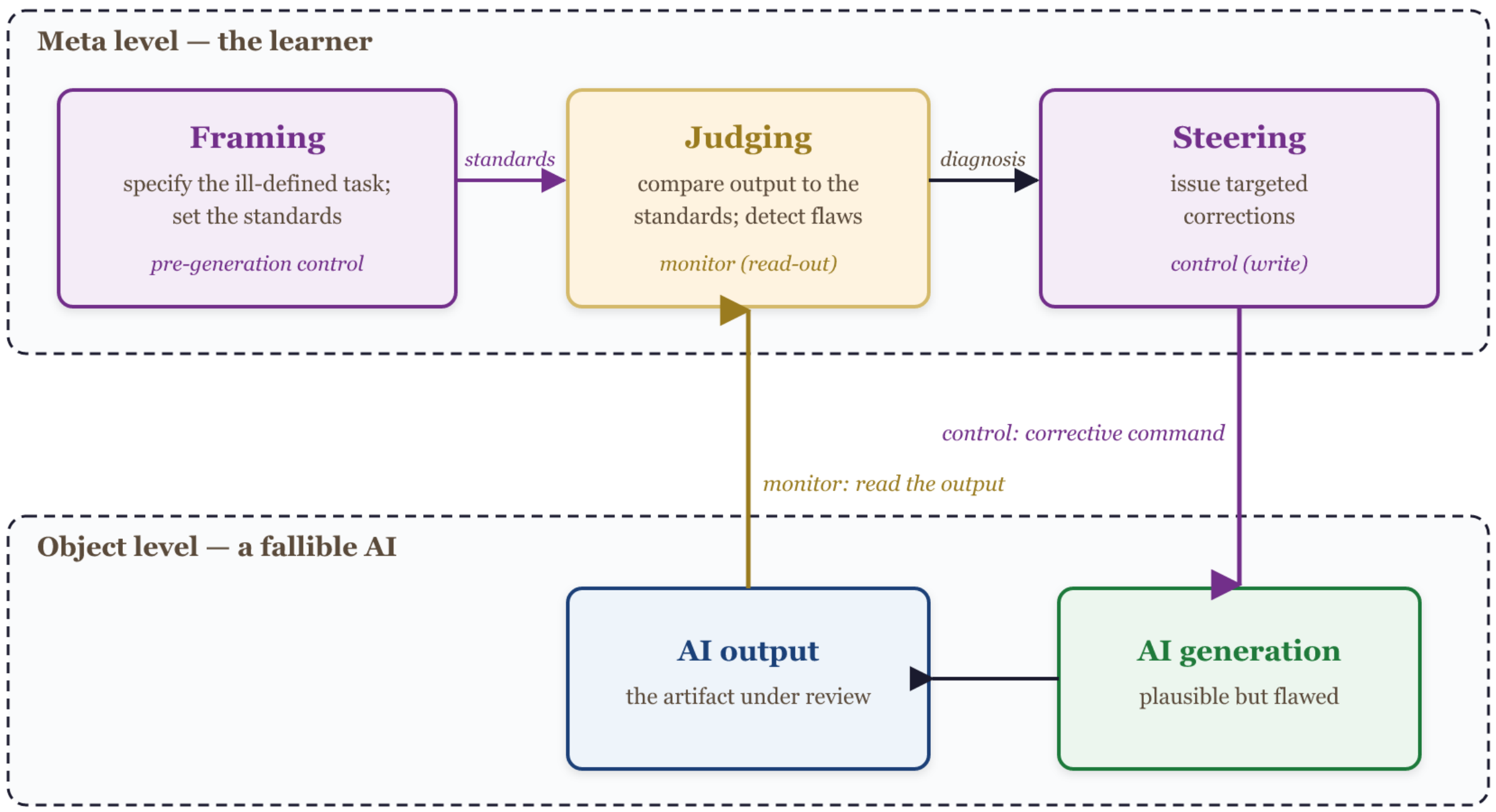

***Figure 2.*** *The CoRe-3 loop. An upstream task definition (Framing, a self-regulated-learning forethought activity) sets the standards against which a metacognitive monitor-control cycle (Judging then Steering) supervises a fallible AI at the object level. Judging is the monitor's read-out; Steering is the controller's write.*

### 3.2 The temporal-separation claim (what makes the decomposition non-obvious)

Existing frameworks treat "use the AI well" as one skill or, at most, pair "prompt" with "evaluate." CoRe-3's distinctive move is to separate two operations that prior models fuse: the *pre-generation* skill of structuring the task (Framing) and the *post-generation* skill of correcting the output (Steering). These are different cognitive acts at different points in time, with different error modes and different instructional remedies. A learner can frame impeccably and steer poorly (detect a flaw but issue a vague correction), or steer fluently atop a malformed task (drive the AI energetically toward the wrong target). Collapsing both into "prompting" hides exactly the distinctions an educator needs.

### 3.3 Distinguishing Judging from Steering

Because Judging and Steering both occur after generation and in a tight loop, their boundary needs stating precisely. **Judging is assessment; Steering is action.** Judging produces a *representation of what is wrong* with the current output, an internal or articulated list of detected flaws, gaps, and risks, and a calibrated sense of how much to trust the output. It is evaluative and its output is a diagnosis. Steering consumes that diagnosis and produces a *corrective instruction* aimed at changing the next output: it is generative and directive, and its quality depends on prioritization (addressing the most critical flaw first), specificity (an actionable command rather than "improve this"), and effectiveness (whether the output actually converges). In the monitor-control terms of Section 3.1, Judging is the monitor's *read-out* and Steering is the controller's *write*. The two dissociate because the competencies differ: a learner may diagnose accurately yet communicate the fix poorly (good Judging, weak Steering), or issue fluent, confident commands that target the wrong thing because the diagnosis was wrong (weak Judging propagating into misdirected Steering, the failure mode Proposition P2 predicts). This last point also reconciles an apparent tension: P2 (Judging bounds Steering) implies the two grades will be *positively correlated* in the aggregate, while P3 claims they *dissociate*. Both hold. P2 is a ceiling relation (Steering cannot exceed the quality its Judging permits), which induces correlation without identity; P3 is the claim that the off-diagonal is well below the reliability ceiling, so the skills are not interchangeable. We therefore test dissociation pairwise and report whether each skill's grade responds to its own manipulated competence while remaining comparatively flat in the others, rather than relying on a single global factor model.

## 4. Theoretical grounding

Each skill is anchored in established theory, and the anchors are mutually consistent because they share the metacognitive backbone of Section 3.1.

**Framing** is the task-definition phase of self-regulated learning. In the COPES model (Winne & Hadwin, 1998), self-regulated work begins by constructing a definition of the task and the standards a product must satisfy; Zimmerman's (2000) forethought phase similarly precedes performance with goal setting and strategic planning. Framing applies this phase to a human-AI loop: the learner converts an ill-defined situation into a specified task with explicit constraints and success criteria. In Bloom's revised taxonomy (Anderson & Krathwohl, 2001), specifying an original task is a Create-level activity, and the competence to know what makes a task tractable for a given tool is metacognitive task knowledge (Flavell, 1979).

**Judging** is metacognitive monitoring (Nelson & Narens, 1990) directed at an external generative source. Its content is the evaluative core of critical thinking, the Delphi-consensus skills of analysis and evaluation (Facione, 1990) and the Paul and Elder (2006) intellectual standards, which supply a ready vocabulary for assessing reasoning. Because the object being judged is a *communicated knowledge claim* from a fluent but fallible source, the most precise anchor is epistemic vigilance (Sperber et al., 2010), which pairs source monitoring (is this source trustworthy?) with content evaluation (is this internally and externally coherent?). Barzilai and Chinn's (2018) account of apt epistemic performance adds the criteria-for-good-knowledge dimension that a learner must hold to judge well, and the human-automation literature supplies the calibration target: reliance matched to actual reliability (Lee & See, 2004), the failure of which is the over-reliance documented in AI-assisted decision-making (Bansal et al., 2021; Buçinca et al., 2021) and the broader automation-complacency it extends (Parasuraman & Manzey, 2010). That Judging is a trainable skill rather than an automatic byproduct of competence is underscored by recent evidence that metacognitive monitoring can decouple from performance in human-AI reasoning (Fernandes et al., 2024), and by recent work on whether learners can evaluate AI output quality as experts do (Nazaretsky et al., 2025).

**Steering** is metacognitive control (Nelson & Narens, 1990): acting on the monitoring signal to change the object-level process. Pedagogically it inverts cognitive-apprenticeship coaching and scaffolding (Collins, Brown & Newman, 1989): the learner, not the master, supplies the corrective guidance. Its quality depends on the learner monitoring the work against held standards (Sadler, 1989), and is well described by feed-forward, the "where to next" component of effective feedback (Hattie & Timperley, 2007).

**The loop and its positioning.** The Judge-Steer cycle is designed to push learners into the Interactive mode of the ICAP framework (interactive, constructive, active, passive; Chi & Wylie, 2014), in which knowledge is co-constructed through dialogue rather than passively received, the mode ICAP associates with the greatest learning. The framework casts the AI as a mediating cultural tool that extends the learner's zone of proximal development (Vygotsky, 1978): the learner accomplishes with the model what they could not alone, while internalizing the Framing, Judging, and Steering moves for eventual independent use.

**The pedagogical stance.** CoRe-3's rejection of speed-to-answer rests on the literature of productive struggle and desirable difficulties (Bjork & Bjork, 2011; Kapur, 2008): conditions that slow performance but deepen learning. The friction of specifying, evaluating, and correcting is not an obstacle to be engineered away but the very locus of learning, which is why the instructional design deliberately presents imperfect output for the learner to improve rather than a polished answer to accept.

## 5. Propositions

- **P1 (Framing gates the loop).** Framing causally precedes and bounds Judging and Steering; framing failures are not recoverable by downstream steering, because a malformed task gives the monitor no stable referent.
- **P2 (Judging gates Steering).** Steering quality is upper-bounded by Judging quality: undetected flaws cannot be corrected, and high steering effort over poor judging yields confident misdirection (the over-reliance failure mode).
- **P3 (Dissociation).** The three skills are separable competencies; proficiency in one does not imply proficiency in another. This is the central empirically-testable claim, and it establishes that the three skills are genuinely distinct rather than a relabeling of "prompting."
- **P4 (Asymmetric transfer).** Judging is bounded by domain knowledge (the calibration trap) and therefore transfers across domains less readily than the more structural skill of Framing.
- **P5 (Inverse of offloading).** Each skill re-inserts a self-regulatory operation that cognitive offloading short-circuits; the framework offloads *execution* while retaining *cognition*.

These propositions differ in how far the present work tests them. The feasibility demonstration (Section 8) directly supports **P3** (the three skills dissociate) and bears partly on **P2** (the steering own-effect is bounded in a way consistent with judging gating steering). **P1**, **P4**, and **P5** are stated here as falsifiable hypotheses for the validation and classroom agenda (Section 9.3), not as claims the demo establishes. Each names a concrete test: P1 fails if a strong steering intervention recovers grades after a deliberately malformed framing; P4 fails if Judging transfers across domains as readily as Framing; P5 fails if exercising the loop does not reduce the offloading signatures (for example, reduced post-task recall) documented in the cognitive-debt literature.

## 6. Relation to prior frameworks

The constructs that compose CoRe-3 are not individually new; what is new is their separation into three parallel, independently-assessable competencies anchored in a monitor-control architecture. We make the boundary explicit by confronting the nearest priors directly.

**AI-fluency frameworks.** The closest practitioner framework is the 4D model of AI fluency (Dakan & Feller, 2025): Delegation, Description, Discernment, Diligence. Its Discernment maps cleanly onto Judging, but its Description bundles two operations we deliberately separate: crafting the initial specification and conducting the iterative back-and-forth. CoRe-3's contention is that these are distinct skills at distinct times, the pre-generation act of Framing and the post-generation act of Steering, with different error modes (a malformed task versus a mis-targeted correction) and different instructional remedies. CoRe-3 also derives its skills from learning theory and an assessment rationale rather than from a fluency heuristic, and so yields rubrics and dissociation predictions that a checklist does not.

**Metacognitive analyses of generative AI.** The strongest construct-level neighbour is Tankelevitch et al. (2024), who analyse generative-AI use through the lens of metacognitive demands, naming prompting, output evaluation, and workflow iteration as sites of metacognitive monitoring and control. We share the metacognitive foundation but differ in goal: their contribution is a demands analysis (a cognitive-load lens that explains why GenAI is hard to use well), whereas ours is an assessable competency model with explicit rubrics, propositions, and feasibility evidence that the components dissociate. The two are complementary: their analysis motivates why each of our skills is cognitively demanding; our framework makes each one measurable.

**Prompt-literacy frameworks.** Prompt-literacy models such as CLEAR (Lo, 2023) and the broader prompt-literacy literature define competence as constructing a precise prompt and iteratively refining it. The most developed recent decomposition, an operationalization of prompt literacy into formulate, interpret, and refine sub-practices (Tour & Zadorozhnyy, 2025), is the closest competitor to our triad; but it bundles task specification and iterative refinement under "prompting" and does not treat the sub-practices as separately graded, dissociable competencies. This fusion of Framing and Steering into a single "prompting" skill is exactly the conflation CoRe-3 rejects. Treating them separately is consequential: it predicts, and our feasibility demonstration supports, that a simulated learner's model-assigned Framing and Steering grades can diverge.

**Metric frameworks for human-AI cognition.** A distinct 2026 line proposes named multi-metric schemes for working with AI, for instance a cognitive-amplification-versus-delegation framework with dependency and drift metrics (Di Santi, 2026). These measure the *sustainability* of reliance rather than teachable framing, judging, and steering competencies, and so are complementary to, not competitive with, an assessable-skill decomposition.

**Knowledge-oriented AI-literacy frameworks.** Field-defining competency sets such as Long & Magerko (2020) and the UNESCO (2024) student framework define AI literacy primarily as understanding AI systems, their capabilities, limits, and ethics, rather than as executing tasks within a human-AI loop. They contain no Framing or Steering construct and only a diffuse notion of critical evaluation that partly overlaps Judging. CoRe-3 is orthogonal: it specifies the task-execution competencies these frameworks leave implicit.

**Empirical accounts of AI-use modes.** Field studies describe how skilled users actually work with AI: the "Cyborg" mode of continuous push-back-and-validate (Randazzo et al., 2025) and the sharp skill-dependence of AI's value at the competence frontier (Dell'Acqua et al., 2023). These describe the behaviour; CoRe-3 supplies the assessable skill decomposition that underlies it.

Table 1 makes the boundary explicit by mapping each nearest prior framework's constructs onto Framing, Judging, and Steering. The recurring pattern is that prior frameworks either (i) omit a construct, or (ii) *fuse* Framing and Steering into one "prompting/iteration" skill.

***Table 1.*** *Where prior frameworks place the three FJS skills (Framing, Judging, Steering).*

| Prior framework | Framing (pre-generation) | Judging | Steering (post-generation) |
|---|---|---|---|
| 4D AI Fluency (Dakan & Feller, 2025) | Delegation + part of Description | Discernment | *fused into* Description |
| Metacognitive demands (Tankelevitch et al., 2024) | "prompting" (as a demand, not a skill) | "evaluating outputs" | "workflow iteration" |
| Prompt literacy / CLEAR (Lo, 2023) | *fused into* "prompting" | weakly present | *fused into* "iterative refinement" |
| AI literacy (Long & Magerko, 2020; UNESCO, 2024) | absent | diffuse "critical evaluation" | absent |

| Prior framework | Framing (pre-generation) | Judging | Steering (post-generation) |
|---|---|---|---|
| Cyborg/Centaur modes (Randazzo et al., 2025) | "directed" mode (described, not assessed) | "push back / validate" | "continuous dialogue" |

No prior column cleanly separates the pre-generation and post-generation skills *and* treats all three as independently scored competencies. That conjunction is the contribution.

**Validated GenAI-competency instruments.** A parallel line of work builds psychometric instruments for AI and generative-AI competence, including validated scales such as GenAIComp (Lee et al., 2025; with factors derived from digital-competence frameworks) and assessment tests such as GLAT (Jin et al., 2024). These establish that GenAI competence can be measured, but their factor structures are literacy-oriented (information literacy, ethics, content creation) and do not isolate a Framing, Judging, or Steering construct. The work closest to ours pairs AI-collaboration literacy with metacognition (Sidra & Mason, 2025) and includes an AI-evaluation sub-construct that overlaps Judging; we differ in separating the pre-generation and post-generation control skills and in demonstrating their dissociation rather than positing correlated factors. CoRe-3 is complementary to this measurement program: it supplies the specific, theory-derived decomposition that a future validated instrument could operationalize. Rubric-based LLM grading studies report high agreement with human raters (intraclass correlation up to 0.97) yet systematic divergence on subjective criteria (Yavuz et al., 2025); the multitrait-multimethod design we use across three grader backends (Section 8) treats that grader-method variance as a measured term rather than a confound.

**Problem formulation as the AI-era skill.** A prominent strand argues that, as models absorb execution, the durable human skill shifts from prompt crafting to *problem formulation*, identifying, analyzing, and delineating the problem worth solving (Acar, 2023). This is precisely our Framing construct, and classroom work has begun to assess it directly, for example through "prompt problems" that require students to specify and evaluate rather than only prompt (Denny et al., 2024), and through instruments that treat question formulation (Kim et al., 2025) and problem decomposition (Srinath et al., 2025) as independently measurable, trainable skills in generative-AI settings. That Framing is a distinct competency is further supported by the older problem-finding literature, which established problem finding as empirically separable from problem solving (Runco & Chand, 1995). We build on this strand by placing Framing in a measured loop with Judging and Steering.

**Evaluative judgement and named-competency frameworks.** The capacity to discern quality, *evaluative judgement*, has been argued to be the core human capability for the generative-AI era (Bearman et al., 2024), and students are observed to make continuous in-the-loop accept-and-reject judgements while working with AI on assessment tasks (Walton et al., 2025). Our Judging skill operationalizes evaluative judgement specifically for AI output; CoRe-3 then departs from a unified-judgement account by separating it from the generative acts of Framing and Steering, treating evaluation as one of three independently-measurable skills rather than a single overarching disposition. Two findings reinforce that Judging is a separable, demanding construct: critical evaluation is the single highest-load activity in AI-assisted writing (Yao & Fan, 2025), and users default to uncritically accepting AI output absent intervention (Wingerter et al., 2025). Among named competency frameworks, the AI Quotient enumerates eight broad collaboration dimensions (Ganuthula & Balaraman, 2025); CoRe-3 instead collapses to three minimal, independently-assessable skills and, crucially, splits the single prompt-engineering dimension that such frameworks leave fused into a pre-generation Framing skill and a post-generation Steering skill. Frameworks that partition coursework by assessment mode rather than by cognitive skill (Elshall & Badir, 2025) are orthogonal and complementary.

**One-line novelty.** CoRe-3 is, to our knowledge, the first theoretically-grounded decomposition of productive generative-AI use into three independently-assessable competencies that separates pre-generation Framing from post-generation Steering, with feasibility evidence that the three skills dissociate. The contribution is the *separation* itself, which is both theoretically motivated (monitor-control plus an upstream task definition) and empirically consequential (the skills can be measured apart).

## 7. The CoReasoningLab system

The framework is instantiated in *CoReasoningLab*, a runnable open-source web platform, which we describe here so that the abstract skills map onto a concrete learner experience. The platform is a Node/Express application with role-based access (student, instructor, administrator), a challenge database, practice and assessment modes, multiple-choice and open-ended response formats per phase, and a five-language content library (English, Hebrew, German, Spanish, French). Its full source, the web application, the database schema, the content library, and the *scoring engine* of sixteen prompts (Section 7.1), is released, alongside a pedagogical-foundations document mapping the design to learning theory. Figures A1 and A2 are screenshots of the running platform, and Appendix A walks through the system end to end. All challenges evaluated in this paper are English-language (Section 9.2), and the quantitative results come from the scoring engine over controlled inputs rather than from production usage logs.

Because challenge generation, rubric generation, and scoring are themselves prompt-defined, the engine extends to a new subject, language, or grader backend by substitution rather than redesign: the feasibility demonstration exercises it unchanged across ten subjects (Section 8) and two grader models, and the released artifact lets others add their own. The platform is therefore **domain-general**: it is not a computing-education tool but a generic instrument for any discipline in which a learner must specify an ill-defined task, judge a fallible solution, and steer it toward a better one. The released content already spans **twelve disciplines** across STEM, the social sciences, the humanities, law, and professional and education fields, from algorithms and classical mechanics to constitutional law, applied ethics, and instructional design (Appendix C). An educator in any of these fields authors a challenge by naming a subject; the system generates the ill-defined problem, the three per-skill rubrics, and the seeded-flaw solution. The monitor-control scaffold is likewise open to additional skills where a setting calls for them; the three are a deliberately small decomposition that separates pre-generation, evaluative, and post-generation control. The released instrument is therefore usable as a starting point for others to extend, not only to reproduce.

**Authoring flow (instructor).** An instructor defines a challenge by choosing a course and subject path; the system then generates the ill-defined problem, the three per-skill rubrics, the gold-standard framing, and the seeded-flaw solution that the learner will critique (Section 7.1). Challenges are organized into courses and can be assigned to cohorts.

**Learner flow (student).** From a dashboard of assigned challenges, a student enters a challenge run that walks through the framework's two phases (Figure A1):

1. *Phase 1: Framing.* The student is shown the raw, ill-defined problem and adds refinement sections (assumptions, constraints, clarifications, success metrics) or selects refinements in multiple-choice mode, then submits. The platform returns rubric-driven Framing feedback and a grade.
2. *AI generation.* The system produces a plausible but deliberately flawed solution to the framed task.
3. *Phase 2: Judge/Steer cycles.* In each cycle the student first **judges** the current output (flagging issues it contains) and then **steers** the AI (issuing correction commands); the AI returns an updated output. The student repeats this up to a configured maximum number of cycles and marks the task complete when satisfied.
4. *Per-skill feedback and grades.* Framing, Judging, and Steering are scored separately, each with its own rubric-driven feedback, and surfaced in a per-challenge report and in longitudinal student analytics that track the three skills independently over time.

This separation in the interface, distinct phases, distinct feedback channels, and distinct grade columns, is the framework's central claim made operational: a learner sees, and is scored on, three different things they did, not one undifferentiated "AI use." The remainder of this section describes the instrument that produces those scores.

### 7.1 Operationalization: the assessment instrument

To show that the three constructs are not only conceptually distinct but practically measurable, we describe a working instrument that scores each skill from a learner's transcript. The instrument is a pipeline of large-language-model prompts; we use it here as an existence proof that automated, rubric-driven scoring of Framing, Judging, and Steering is feasible, not as a validated assessment.

**Challenge construction.** Each challenge begins with a deliberately ill-defined problem generated to contain two or three unstated gaps, recorded internally but never shown. A per-challenge set of three rubrics, one each for Framing, Judging, and Steering, is generated for the subject area, each with three to five measurable criteria and explicit excellent and poor indicators. A gold-standard "best framing" is generated as an internal reference. The design instantiates an inverted cognitive apprenticeship: rather than observing an expert, the learner is given a fallible artifact to repair. This connects the instrument to the instructional literature on learning from errors and erroneous examples, in which studying and correcting flawed solutions improves error detection and conceptual understanding relative to studying only correct ones (Große & Renkl, 2007; Durkin & Rittle-Johnson, 2012); CoRe-3 generalizes that paradigm from static worked examples to an interactive, learner-driven repair loop over AI output.

**The deliberately-imperfect output.** After the learner frames the task, the model produces a plausible, professional-looking solution that is required to embed two to four non-trivial issues, wrong-but-reasonable assumptions, missing edge cases, or subtle logical errors, each recorded internally with a severity label and none flagged to the learner. Across steering cycles, updates address the learner's commands but may introduce new minor issues, so that difficulty adapts to the quality of steering rather than collapsing to a perfect answer.

**Scoring.** Each skill is scored in two stages. A skill-specific evaluator assesses the learner's response against the (internal) rubric and produces per-criterion ratings on a three-point scale; a generic grading stage then aggregates those ratings into a final grade, weighting critical criteria more heavily rather than averaging. Crucially, the three evaluators differ in what they compare against: Framing is evaluated against the gold framing and the rubric; Judging is evaluated against the seeded ground-truth issues, yielding a recall/precision signal (issues correctly identified, missed, and falsely flagged); and Steering is evaluated against the trajectory of the output across cycles, rewarding corrections that demonstrably move the solution toward correctness. This is why the skills are measured *apart*: each evaluator interrogates a different referent.

The instrument used in this paper is the scoring engine of the CoReasoningLab platform, a library of sixteen prompts spanning challenge construction, AI generation, and evaluation (Figure A3 traces the full generation-and-assessment pipeline). The released library carries the platform's evaluation logic verbatim, the same rubrics, criteria, and system prompts as the deployed application; the only adaptation is that JSON output formatting is supplied by the research harness rather than embedded in each prompt. The surrounding harness (simulated-learner generation, the crossed factorial, and the analysis) is research orchestration around that fixed evaluation logic, not a reimplementation of it, so the measurements characterize the instrument itself. Scoring a single learner exercises the relevant subset, the three skill evaluators and the generic grader, over controlled inputs, so that the measurements are reproducible and the ground truth is known. A scope note: the grader is a language model, so the results below characterize the *internal* behavior of the instrument (whether it separates controlled competence levels and dissociates the skills); agreement with human experts is the separate validity question the prepared study in Section 9.3 addresses.

## 8. Feasibility demonstration

We exercise the instrument over controlled inputs to test three feasibility claims: that it *discriminates* competence, that the three skills *dissociate* (Proposition P3), and that the grader is *self-consistent* enough to report. The "learners" are simulated personas of controlled per-skill competence, generated by one model (gpt-4o-mini) and graded by a *different* model (gpt-4o), so that no model grades its own output. These claims concern the instrument; the agreement study with human expert graders is prepared and is the next stage (Section 9.3).

**Design.** We use a crossed factorial: each of the three skills is independently set to a strong or weak competence level, giving $2^3 = 8$ profiles, crossed with ten challenges across ten distinct subjects (algorithms, microeconomics, machine learning, databases, statistics, operating systems, calculus, organic chemistry, linguistics, and corporate finance), for 80 simulated learners. Framing and Steering responses are generated by the competence-conditioned learner model; Judging is operationalized by a competence-conditioned selection over the challenge's ground-truth seeded issues (a strong judge flags all real issues and no false ones; a weak judge flags few real issues and some false ones). These seeded issues and the distractors are themselves machine-generated and not yet human-verified, so Judging's by-construction own-effect is explicitly contingent on that ground truth being correct. Because the manipulation is per skill, the design can separate whether each grade tracks its *own* skill's competence (discrimination) from whether it is *insensitive* to the other skills' competence (dissociation).

**Discrimination.** Grades move monotonically with competence: the all-weak profile averages C on every skill, the all-strong profile averages between B and A, and each skill's grade rises when that skill is set to strong. The judging signal is mechanistically transparent: a strong judge flags all of the seeded issues with no false alarms and is graded A; a weak judge flags none and raises false issues and is graded C.

**Dissociation (the central result).** Table 2 reports, for each graded skill, the change in its mean grade (on a 3-point scale, A=3..C=1) when each skill in turn is moved from weak to strong. The result that carries the claim is the two *blind-graded* skills, Framing and Steering, whose free-text responses the grader scores without knowing the intended competence: each shows a clear positive own-effect (**+0.62** and **+0.43**) with near-zero cross-effects, so a simulated learner's two free-text grades move independently. Judging's own-effect (**+2.00**) is large but *fixed by construction*: its competence is operationalized by a controlled selection over the challenge's seeded issues, and those seeded issues are themselves machine-generated and not yet human-verified (Section 9.2). We therefore read Judging's clean diagonal as a controlled check that the grader correctly rewards recall and precision, not as independent evidence that the skills separate. The off-diagonal (cross-skill) effects average +0.01 across the matrix, so each grade responds to its own skill and is essentially flat in the others (Figure 3).

***Table 2.*** *Effect on each skill's grade of manipulating each skill's competence (grade Δ, strong − weak; N=80).*

| grade of ↓ \ manipulated → | Framing | Judging | Steering |
|---|---|---|---|
| **Framing** | **+0.62** | −0.02 | −0.07 |
| **Judging** | +0.00 | **+2.00** | +0.00 |
| **Steering** | −0.12 | +0.27 | **+0.43** |

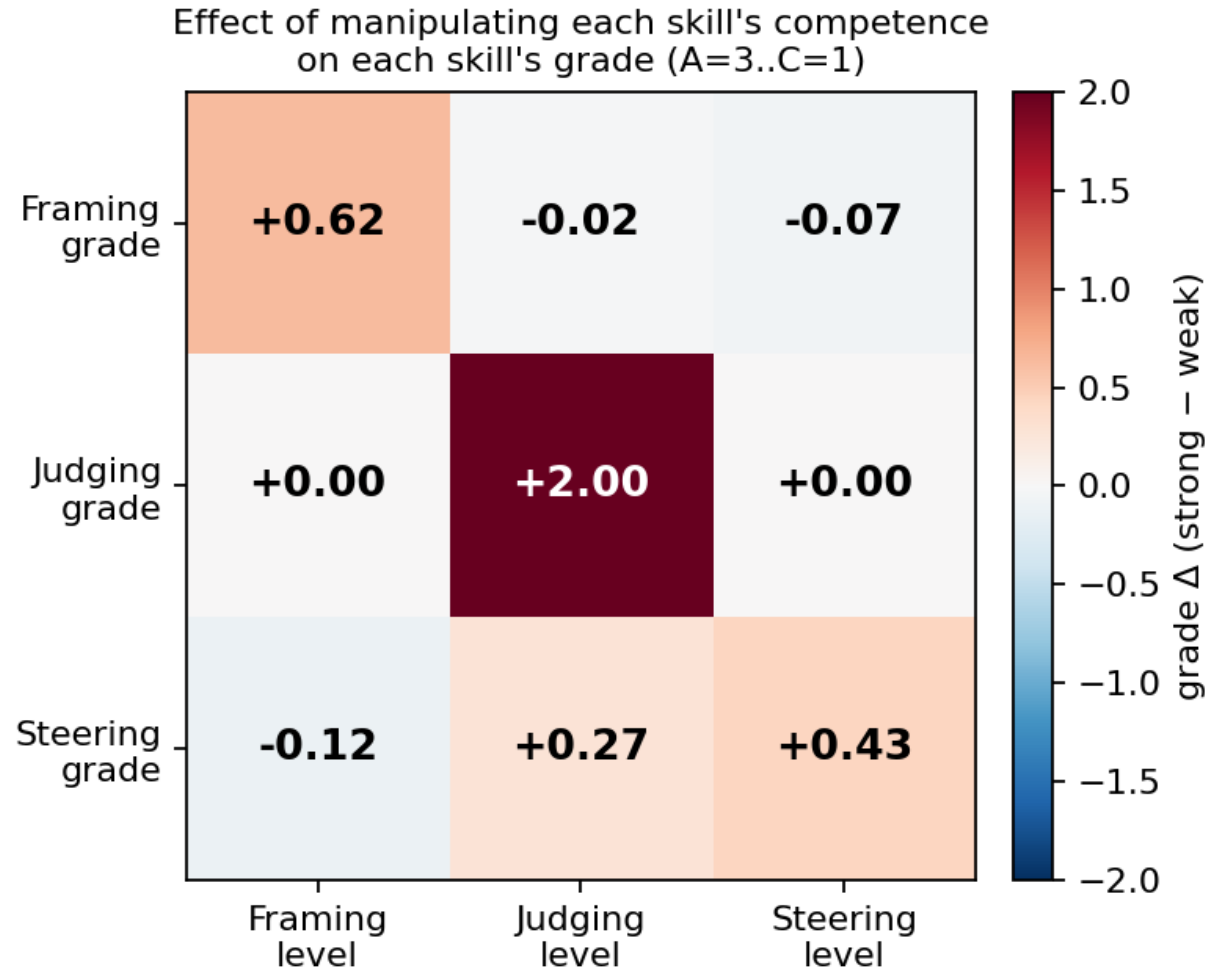

***Figure 3.*** *Effect of manipulating each skill's competence (columns) on each skill's grade (rows). Each cell is the grade change (strong minus weak) on the 3-point scale (A=3, B=2, C=1); warmer cells are larger positive effects. The diagonal (own-skill effect) dominates; off-diagonal (cross-skill) effects are near zero.*

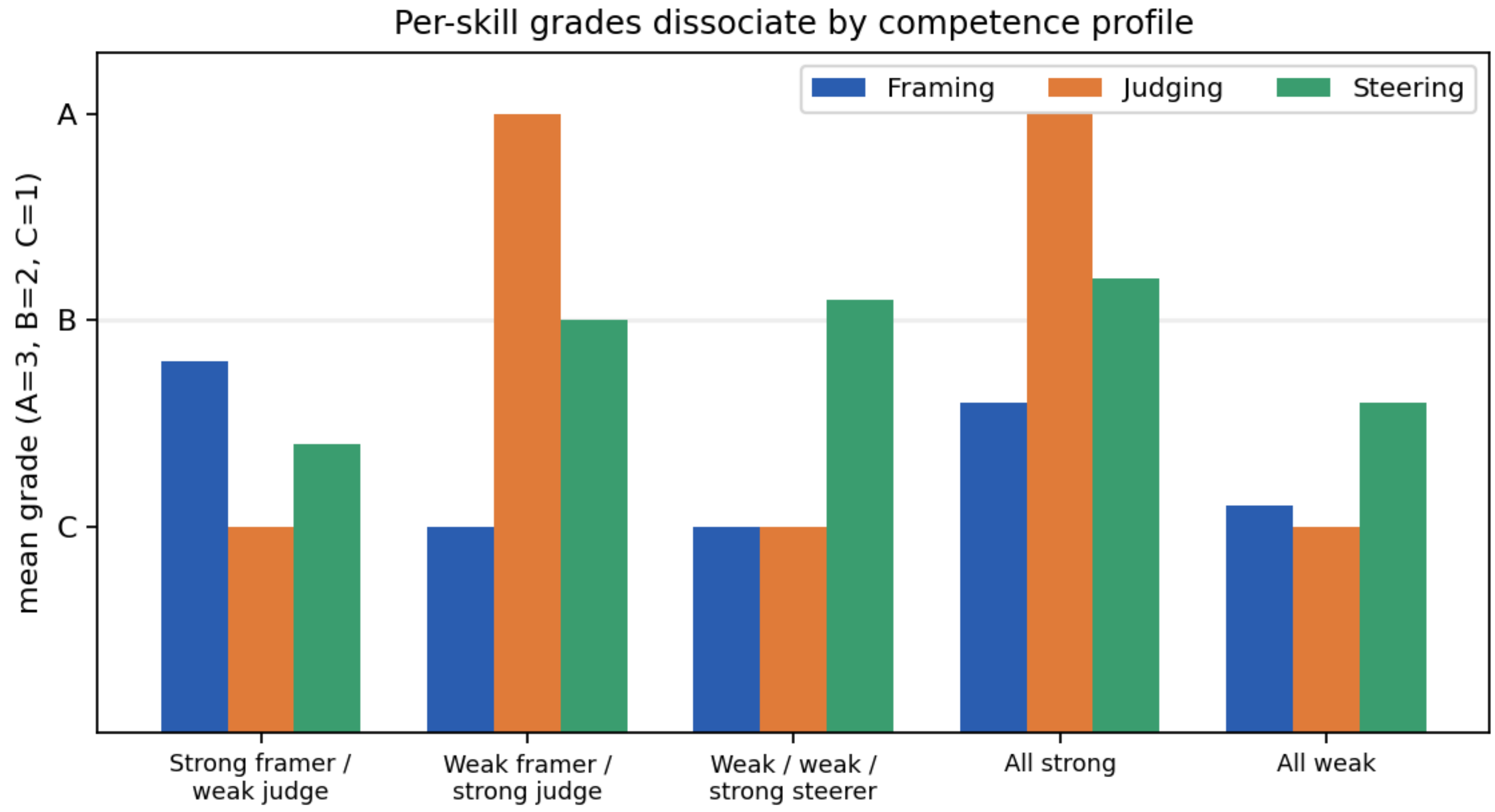

***Figure 4.*** *Mean per-skill grade for five competence profiles. Judging reaches A only when judging is strong, regardless of framing or steering; each skill responds to its own competence.*

The diagonal (own-skill) effects are statistically reliable: bootstrap 95% confidence intervals (2,000 resamples) exclude zero for Framing (+0.62, CI [+0.46, +0.77]) and Steering (+0.43, CI [+0.20, +0.65]); Judging is deterministic by construction (+2.00). The same separation appears in the inter-skill grade correlations across the 80 learners: Framing-Judging $\rho = -0.03$ ($p = 0.82$) and Framing-Steering $\rho = -0.12$ ($p = 0.29$) are both non-significant, while Judging-Steering $\rho = +0.25$ ($p = 0.02$) is positive and significant. The **Framing-Judging** pair is the decisive demonstration: these two skills are scored by entirely separate mechanisms (free-text framing evaluation versus issue-selection judging) yet their grades are uncorrelated, which a single general-ability account cannot produce. The Judging-Steering correlation behaves exactly as Proposition P2 predicts, the expected ceiling relation in which judging gates steering, and so is consistent with, rather than evidence against, separability. With only three indicators a formal factor model is under-identified, so we rest the separability evidence on the manipulation-based effect matrix and the inter-skill correlations rather than on a confirmatory dimensionality test.

**Convergent and discriminant validity (a multitrait-multimethod view).** The result so far is a *discriminant*-validity finding: three skills (traits) separate. Construct validity also requires the complementary *convergent* property, that the same skill measured by different methods agrees, and that the instrument reports correlation when the underlying competence really is shared (Campbell & Fiske, 1959). Two analyses supply it. First, treating the three grader backends as independent methods, the same skill graded by gpt-4o, gpt-4o-mini, and the independent-vendor llama-3.3-70b agrees at the per-learner level for the two skills with the widest grade range (mean pairwise rank correlation +0.57 for Framing and +0.67 for Judging across the 40 shared transcripts), while *different* skills stay near zero, the convergent-high / discriminant-low pattern a valid measure should show; Steering's narrower range under the P2 ceiling yields a lower cross-method correlation (+0.21), a restriction of variance rather than a separate construct. Laid out as the full multitrait-multimethod matrix, the convergent correlations (same skill across graders, mean +0.49) exceed both the heterotrait-monomethod correlations (different skills within one grader, mean +0.08) and the heterotrait-heteromethod correlations (different skills across different graders, mean +0.06); no grader shows halo (within-grader cross-skill mean $|\rho|$ of 0.13, 0.18, 0.13 for the three backends), so all three Campbell-Fiske criteria hold. Second, the instrument is faithful to the population's dependence structure: on learners whose three competences are set *independently* (the full crossed design) the inter-skill grade correlations are near zero (mean $|\rho| = 0.13$; Framing-Judging $\rho = -0.03$), whereas on the 20 learners whose three competences are all set strong or all set weak (a single shared ability level) the same instrument returns *positive* correlations (mean $\rho = +0.41$; Framing-Judging $\rho = +0.52$). The instrument therefore does not manufacture dissociation: it recovers independence when the competences are independent and correlation when they are shared, which is what licenses reading the crossed-design separation as a property of the skills rather than of the scoring (Table 3).

***Table 3.*** *Multitrait-multimethod validity. Panel A: convergent agreement of the same skill across the three grader backends (mean pairwise per-learner rank correlation on the 40 shared transcripts). Panel B: inter-skill grade correlation by population, showing the instrument recovers the population's dependence structure (discriminant on the full crossed design, convergent on the n=20 all-strong and all-weak learners whose three competences share one level).*

| Panel A: cross-grader convergent ($\rho$, 3 backends) | Framing | Judging | Steering |
|---|---|---|---|
| same skill across gpt-4o, gpt-4o-mini, llama-3.3-70b | **+0.57** | **+0.67** | +0.21 |

| Panel B: inter-skill $\rho$ by population | F-J | F-S | J-S | summary |
|---|---|---|---|---|
| independent (crossed) competences | −0.03 | −0.12 | +0.25 | mean $|\rho|$ = 0.13 |
| shared single-ability level | **+0.52** | +0.14 | **+0.56** | mean $\rho$ = +0.41 |

The own-skill effects are directionally consistent across subjects: broken down by the ten subject areas (eight learners each), Framing and Steering each show a positive own-competence effect in nine of ten subjects, with the two exceptions (Framing in statistics, Steering in microeconomics) reflecting the small per-subject sample rather than a sign reversal; Judging is fixed by construction in every subject. The own-competence effect is therefore not driven by any single domain; the pooled own-effects and their confidence intervals above carry the estimate, and the per-subject pattern shows it holds broadly across the curriculum. The designed-contrast personas make the separation concrete (Figure 4, which plots five profiles). Three are especially telling: a *weak-framer / strong-judge* learner scores Framing C but Judging A; a *strong-framer / weak-judge* learner inverts this to Framing B, Judging C; and a *weak / weak / strong-steerer* elevates only Steering. A single underlying "AI-use ability" cannot produce these crossed profiles. Converging evidence comes from outside our synthetic setting. In an intervention study, students' *behavioral* regulation of LLM use (reformulating queries, checking correctness) predicts effective use, whereas self-rated AI expertise does not (Clerc et al., 2026). The skill of working with AI is thus distinct from a general, self-assessed competence.

**Robustness and ablations.** Three checks, reported in full in Appendix B, support the result. First, the grader is **92% self-consistent** on repeat, a precision check, not accuracy against humans. Second, the dissociation replicates across three grader backends spanning two providers (gpt-4o, gpt-4o-mini, and Meta's llama-3.3-70b; diagonal-to-off-diagonal ratio 39), so it does not hinge on one model snapshot. Third, a ground-truth ablation shows Framing and Steering are scored by rubric-guided model judgment while Judging, as instrumented, tracks the seeded answer key, which locates the Judging score to settings with known ground truth (Section 9.3 specifies the open-ended variant).

**Scope.** These results establish feasibility. (i) Judging's clean diagonal (+2.00, zero cross-effects) is a controlled check by design: its competence is set by a controlled selection over the seeded issues, so the diagonal confirms that the grader correctly rewards recall and precision, and the independent separation evidence is carried by the free-text, blind-graded skills, Framing and Steering, whose positive own-effects (+0.62, +0.43) with near-zero cross-effects drive the result. (ii) The dissociation replicates across grader backends spanning two providers (Appendix B); agreement with human experts is the validity step the prepared study in Section 9.3 supplies. (iii) With three constructs the separability evidence rests on the manipulation-based effect matrix and the inter-skill correlations, the appropriate tests for the design. (iv) Steering's own-effect (+0.43 at N=80) is governed by the P2 ceiling by design: steering quality is bounded by the judging that precedes it, so manipulating steering alone has bounded headroom. This is a property of the construct, not of the grader: under a deliberately strict steering rubric (a C for vague or non-prioritized commands) the steering own-effect is unchanged (+0.40 versus +0.50, statistically indistinguishable) and the dissociation persists (ratio 27.5). Human steering data will trace this ceiling directly.

## 9. Discussion

The feasibility demonstration establishes that the three skills are separable and automatically scoreable; this section reconciles the tensions the framework raises, bounds the present evidence, and specifies the validation program it invites.

### 9.1 Tensions the framework reconciles

Four tensions bear on the framework, each with a resolution within it.

*Offloading versus productive struggle.* Evidence that AI use can depress critical thinking through cognitive offloading (Gerlich, 2025; Kosmyna et al., 2025) appears to threaten any framework that puts learners in close partnership with AI. The resolution is in the zone-of-proximal-development stance: CoRe-3 treats the AI as a mediating tool whose purpose is the learner's eventual *independence*, and it offloads execution while deliberately retaining the cognitive work of specifying, evaluating, and correcting. The framework is, in this sense, the designed inverse of metacognitive laziness (Proposition P5). Whether exercising the loop actually reduces the offloading signatures is an empirical question this paper does not test; it is reserved for the efficacy study of Section 9.3.

*The calibration trap.* Judging is metacognitive monitoring, and monitoring is only useful when calibrated. A learner who lacks the domain knowledge to recognize an error cannot detect that error in AI output, however vigilant. Judging is therefore bounded by domain competence, and the framework should be read as describing a skill that develops *with* domain knowledge, not as a substitute for it (Proposition P4).

*Interactive is not automatically productive.* ICAP predicts that interactive engagement yields the most learning, but rapid AI dialogue can be voluminous and shallow, a sequence of re-rolls rather than reasoning. Steering counts as genuine metacognitive control only when corrections are knowledge-generating, which is why the instrument rewards targeted, convergent corrections rather than mere repetition.

*Standards can regress.* Judging and Steering presuppose that the learner holds standards adequate to evaluate the output. When the AI is more competent in the domain than the learner, the standards the learner applies may be inferior to the artifact under review, a reversal that classical formative-assessment theory does not anticipate and that bounds the framework's applicability at the expert frontier.

### 9.2 Scope and boundary conditions

This is a conceptual contribution with a working instrument, and three boundary conditions define where the present evidence applies. The feasibility demonstration uses simulated learners of controlled competence: they establish that the grader carries signal and that the skills separate, and they set up the human-learner studies of Section 9.3. To keep generation and grading independent, learners are generated by one model (gpt-4o-mini) and graded by another (gpt-4o), and the dissociation replicates across grader backends spanning two providers (Section 8); broadening the grader pool further is part of the validation program. The challenges are English-language, and agreement with human experts is the defined next step (Section 9.3). Two construct-specific boundaries are worth stating. First, Judging as instrumented here is scored against a seeded answer key, so the Judging score applies to settings with known ground truth; open-ended scoring without a key is a distinct instrumentation that Section 9.3 specifies. Second, Steering's signal is governed by the P2 ceiling by design (its quality is capped by the judging it follows), a relation that human steering data will trace directly. What the paper establishes is a theoretically-grounded decomposition with stated propositions, a precise novelty boundary, and evidence that the three constructs are separable and automatically scoreable; the efficacy study of learning outcomes is the next stage of the program.

### 9.3 A validation and assessment agenda

The feasibility demonstration shows that the constructs are separable and measurable; it does not establish that the automated grades match expert human judgment, nor that exercising the skills improves learning. We therefore specify the validation program the framework invites, organized as an argument-based validity case in the sense of Messick (1995) and Kane (2013): the present evidence supports the *scoring* and *generalization* inferences (the instrument scores consistently and the three constructs separate), while the *extrapolation* inference (that the scores reflect a human competency that transfers) and the *implication* inference (that the scores support instructional decisions) remain to be established by the studies below.

First, an instrument-validity study, built and ready to run. Three blind expert raters re-grade a stratified sample of 40 transcripts (covering all eight competence profiles) on the three skills using the per-skill rubrics; inter-rater reliability is reported as ordinal Krippendorff's $\alpha$ (primary, since the grades are ordinal), Fleiss' $\kappa$, and pairwise Cohen's $\kappa$, and agreement of the human majority with the automated grade as Cohen's $\kappa$ and ordinal $\alpha$ per skill, against the field-typical bar of 0.6 to 0.8. With all three raters scoring the same 40 items the agreement estimate has a bootstrap half-width on the order of $\pm 0.1$, enough to place each skill inside or outside that band. This step is indispensable rather than a formality: recent work shows that LLM-as-judge agreement with human experts is moderate and task-dependent, sometimes falling to Fleiss' $\kappa$ near 0.1 to 0.3 on hard rubric judgments (Feng et al., 2025), so automated grades are validated against humans rather than assumed reliable. The package confronts the recursive calibration threat (who grades the grader) on a second level: because the Judging construct rests on machine-seeded ground-truth issues, a parallel task has the same three raters verify a balanced set of 40 seeded issues (20 real and 20 distractor controls) as genuine flaws, yielding a confirmation rate for the Judging ground truth before its recall/precision signal is trusted. The full package, codebook, per-rater shuffled task files with hidden ground truth, and seed-fixed agreement-scoring scripts, is released and reproducible, so the study runs directly once raters are recruited. Second, a construct-validity study at scale (target of at least 200 real learners) to test Propositions P1 through P4, examining whether Framing, Judging, and Steering dissociate across a learner population and whether the proposed gating relations hold. Third, a grader-robustness study across multiple model backends to separate the framework's signal from any single model's idiosyncrasies. Only an efficacy study with real learners can test Proposition P5 and any learning claim; that is explicitly outside the present scope.

Beyond validity, the framework opens a research program, and one strand can begin before the human studies: external validity can be probed by scoring a released corpus of real student-LLM dialogue, the StudyChat dataset of 16,851 student-ChatGPT interactions from a university course (McNichols et al., 2025), with the per-skill instrument, testing whether Framing, Judging, and Steering are detectable and separable outside the controlled challenge format; StudyChat's own dialogue-act labels (questioning, verification, and editing) align with the three skills and supply an external anchor. Four further directions follow from the propositions: a longitudinal study of whether the three skills develop and persist with practice; an instructional-intervention study testing whether targeted teaching of each skill improves it, building on evidence that students' behavioral regulation of LLM use is trainable (Clerc et al., 2026); a transfer study testing P4's prediction that Framing, the more structural skill, transfers across domains more readily than the domain-bound Judging, a question made tractable by evidence that metacognitive regulation can show far transfer (Wirth et al., 2025); and a mediation study testing whether metacognitive monitoring accuracy predicts Judging, extending findings that calibration predicts subsequent strategy use in computer-based learning (Lee & Bosch, 2025) and tying the construct to its monitor-control grounding.

## 10. Conclusion

The task of education in the age of generative AI is not to produce faster answer-getters but to cultivate critical collaborators: learners who can specify a problem worth solving, judge what a machine returns, and steer it toward something better. That capability is teachable only if it can be named and assessed. CoRe-3 offers a decomposition of it into three theoretically-grounded, independently-assessable skills, Framing, Judging, and Steering, separates the pre-generation skill from the post-generation one in a way prior frameworks do not, and shows that the three can be measured apart. Because the decomposition tracks what a person must contribute rather than what today's models cannot yet do, it does not lapse as the models improve: the more capable the system, the more the assessable skill is the framing, judging, and steering of it. We release the model, the instrument, and the validation protocol as a foundation to build on: an open, domain-general platform with which an educator in any discipline, from algorithms to applied ethics, can assess and train this skill, for the assessment and instruction the moment demands.

## Appendix A. System walkthrough

Figure A1 shows a single challenge run in CoReasoningLab, illustrating how the three skills appear as distinct, separately-scored stages of one continuous task. In Phase 1 the learner refines an ill-defined problem and receives a Framing grade. The system then produces a deliberately flawed solution. In Phase 2 the learner judges that output (flagging real issues while avoiding distractors) and steers the AI with a targeted correction; the output converges across cycles. The final report returns three independent grades with per-skill diagnostic feedback, the interface-level expression of the framework's central claim that productive AI use is not one skill but three.

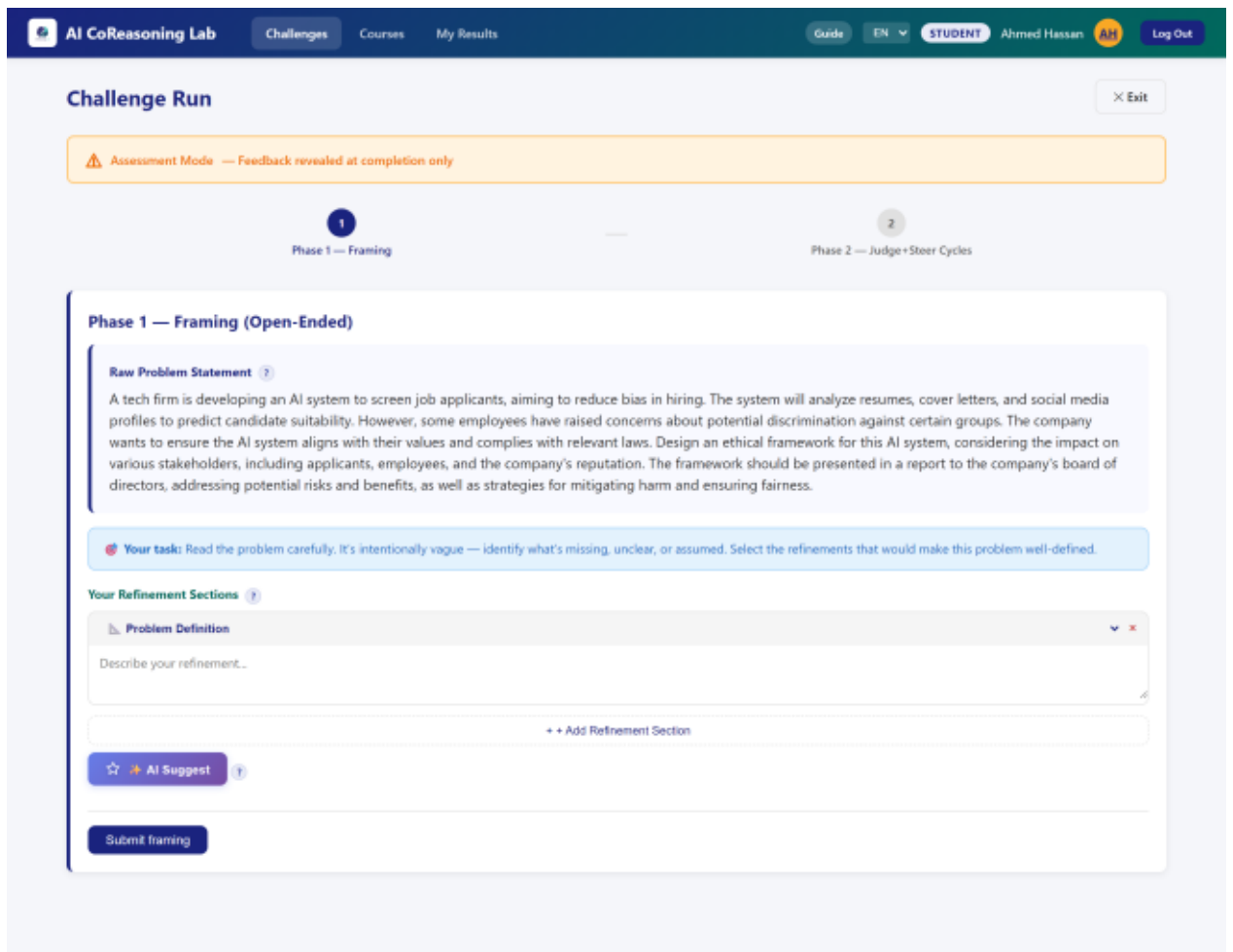

***Figure A1.*** *Screenshot of the running platform: a learner's challenge run at the Framing phase. The engine has generated an ill-defined problem (here, an AI hiring-screening task), and the learner specifies refinements before any solution is produced. The two-phase structure is explicit in the interface, Phase 1 (Framing) and Phase 2 (the Judge-Steer cycle), and each skill is scored and given feedback independently.*

The platform exposes role-specific interfaces beyond the challenge run (Figure A2). Students see a dashboard of pending challenges and courses and a personal results view that trends Framing, Judging, and Steering separately over time. Instructors author challenges (the system auto-generates the ill-defined problem and the three per-skill rubrics) and read course analytics that break grade distributions down by rubric and by student. In every view the three skills remain distinct columns, which is the design commitment the framework makes visible.

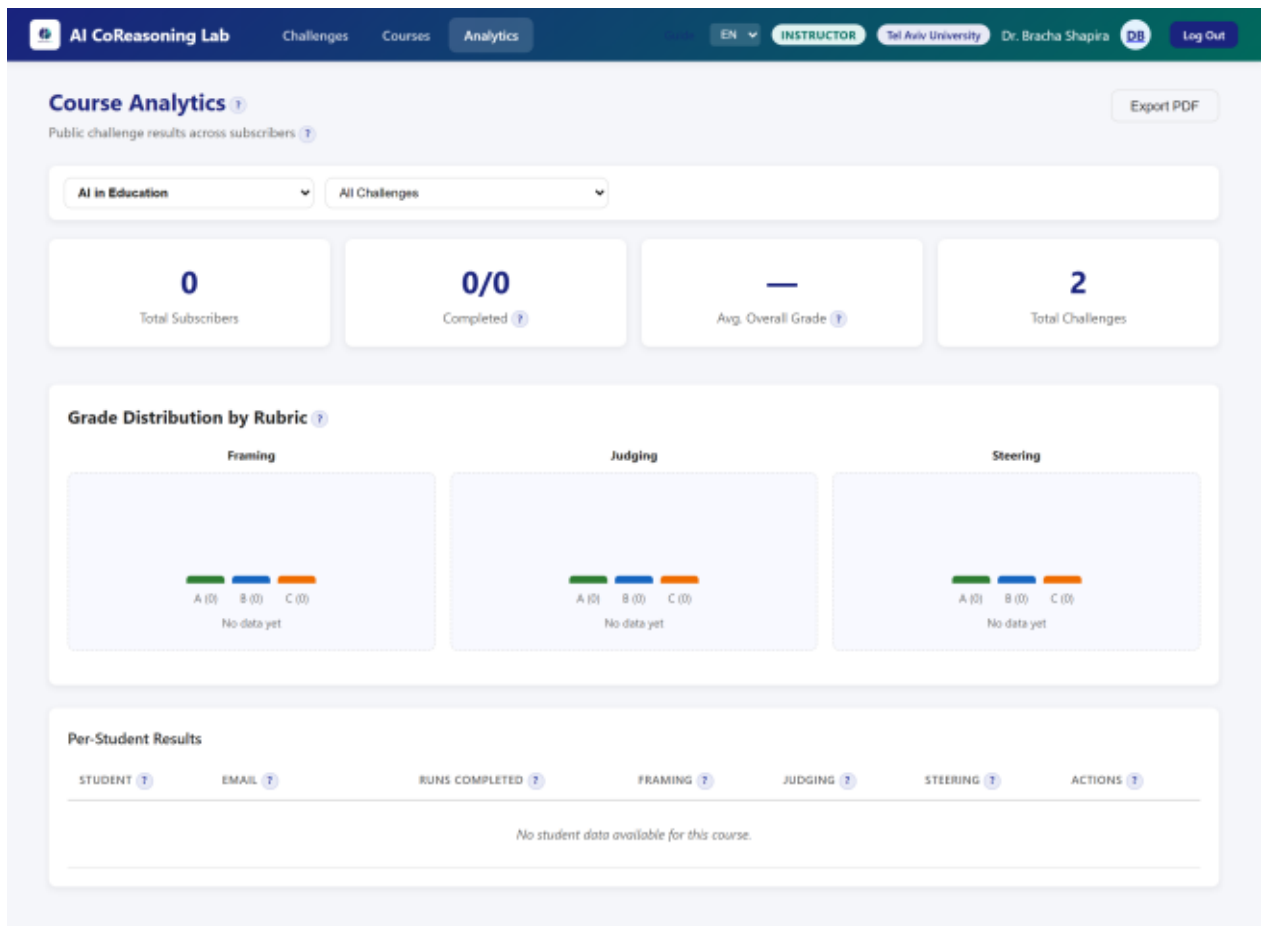

***Figure A2.*** *Screenshot of the running platform: the instructor course-analytics view. Grade distribution is broken out into separate Framing, Judging, and Steering panels, and the per-student results list the three skills as distinct columns, the interface-level expression of the framework's central claim that productive AI use is not one skill but three.*

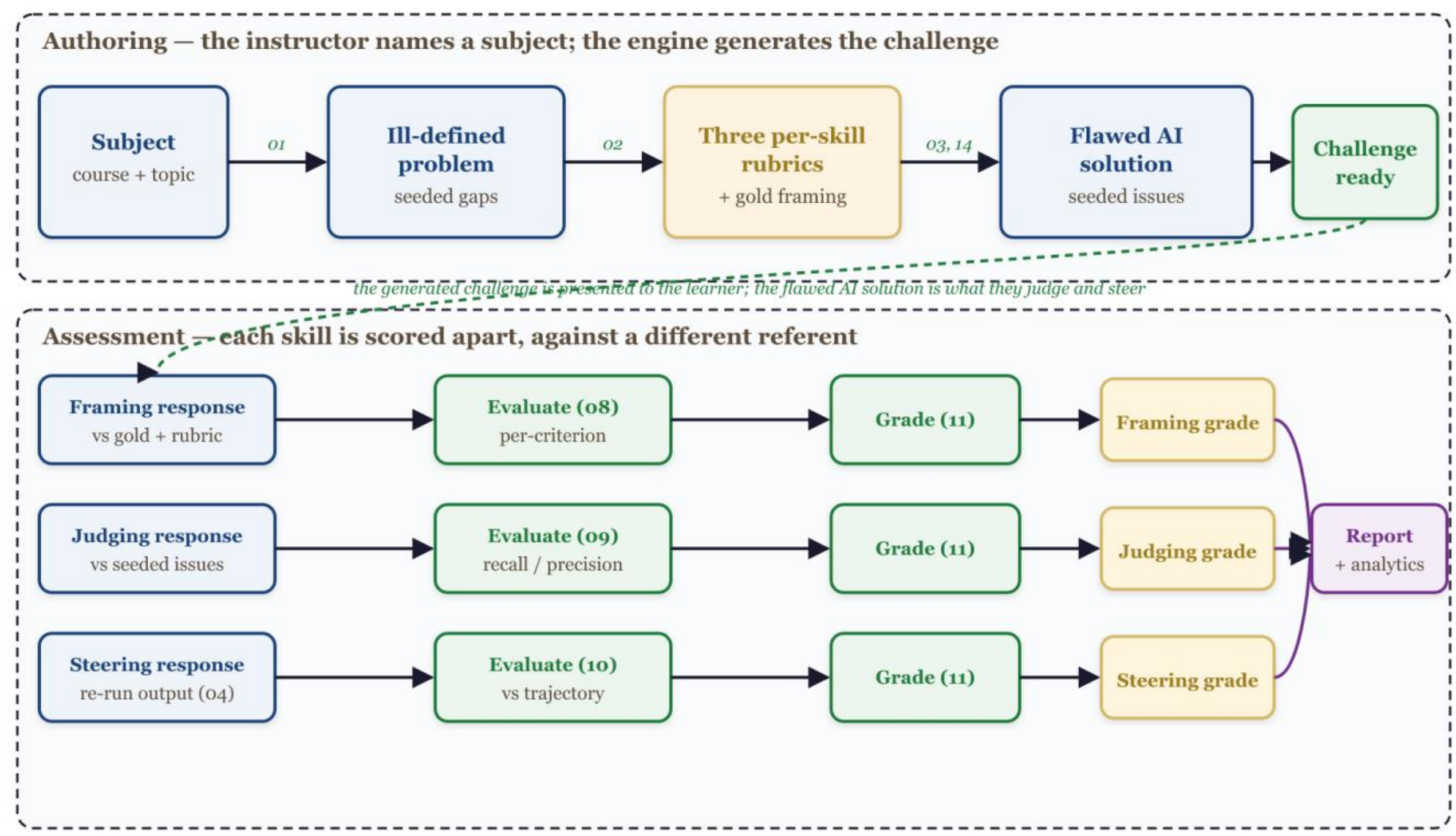

***Figure A3.*** *The sixteen-prompt engine in two phases. In authoring, the instructor names a subject and the engine generates the ill-defined problem (prompt 01), three per-skill rubrics and a gold framing (02, 14), and a deliberately flawed AI solution carrying seeded issues (03). In assessment, each learner response is scored against a different referent: Framing against the gold framing and rubric (08), Judging against the seeded issues (09), and Steering against the corrected trajectory after the output is re-run (04, 10); a single grader prompt (11) then assigns each skill its own grade, and the three feed the per-student report and course analytics. Because the three evaluation prompts interrogate different referents, the grades are produced independently rather than as one global impression.*

## Appendix B. Robustness and ablations

**Reliability.** To check that the grades are not noise, we hold four learner transcripts fixed and re-run the evaluation-and-grading prompts five times each, an independent sample each time at the production temperatures. The grader is **92% self-consistent** overall: the mean grade-flip rate across repeats is 0.08, with Judging deterministic (0.00, by construction), Framing 0.15, and Steering 0.10, and a mean within-cell grade standard deviation near 0.2 on the three-point scale, indicating that disagreements are occasional one-level boundary jitter rather than unstable scoring. This addresses the documented self-inconsistency of LLM judges, though it measures *precision* (repeatability), not *accuracy* against humans (Section 9.3). Because Judging contributes a flip rate of zero by construction, the load-bearing figure is the blind-graded skills, whose flip rate is near 0.12.

**Grader-backend robustness (across vendors).** The dissociation holds across three grader backends spanning two providers. The primary grades come from gpt-4o; re-grading the identical 40 transcripts (challenges and learner responses held fixed) with gpt-4o-mini reproduces it (diagonal +0.47 versus off-diagonal +0.07), and re-grading the same transcripts with an independent-vendor model, Meta's llama-3.3-70b (the prototype's own engine), reproduces it again with a stronger margin: own-skill effects of +0.80 (Framing), +0.75 (Judging), and +0.40 (Steering) against a mean cross-skill effect of +0.02, a diagonal-to-off-diagonal ratio of 39. Two facts stand out across backends. First, Judging's own-effect is +2.00 only under gpt-4o's strict adherence to the seeded-issue selection; under both gpt-4o-mini and the independent-vendor llama it settles to +0.65 and +0.75, level with Framing (+0.65 and +0.80), so the separation is balanced across all three skills, not propped up by the built-in Judging diagonal. Second, every backend reproduces the designed contrast profiles (a strong-framer / weak-judge learner scores high Framing and low Judging, and the inverse for a weak-framer / strong-judge learner), which a single general-ability account cannot produce. Because the result holds across three independently-versioned models from two vendors (gpt-4o, gpt-4o-mini, and Meta's llama-3.3-70b), it does not hinge on a single model snapshot; the released harness pins the model identifiers so the run reproduces, and the cross-backend agreement is the evidence that it generalizes beyond any one of them. Cross-vendor replication is therefore established, not deferred.

**Dependence on ground-truth scaffolding.** To probe how much the instrument's discrimination relies on the seeded ground truth (the gold framing supplied to the framing evaluator and the known seeded issues supplied to the judging evaluator) versus the model's own judgment, we re-grade a 40-learner subset (five subjects) with that scaffolding removed, comparing against that subset's own baseline (Framing +0.60, Judging +2.00, Steering +0.50). The effect is sharply skill-specific. Framing discrimination is essentially unaffected (own-effect +0.75 without the gold reference, versus +0.60 with it): the rubric and the model's own judgment carry it. Judging discrimination, by contrast, falls sharply (+2.00 to +0.45): without the known issues the grader has no recall-precision anchor and no longer separates strong from weak judging. Steering falls in between (+0.50 to +0.25). The skills still dissociate (ratio 29), and the result pinpoints what each grade measures: Framing and Steering are scored largely by rubric-guided model judgment, whereas Judging *as instrumented here* tracks agreement with a known answer key. This both explains Judging's by-construction +2.00 and locates the Judging score precisely: it applies where ground truth is known, and open-ended judging without a key is the distinct instrumentation Section 9.3 specifies.

## Appendix C. Cross-disciplinary challenge showcase

The framework and the platform are domain-general. An ill-defined problem with seeded flaws, three per-skill rubrics, and a deliberately-imperfect AI solution can be generated for any subject, because each is produced by a prompt that an educator parameterizes with a course and topic. The released platform ships challenge content across twelve disciplines (Table C1) spanning STEM, the social sciences, the humanities, law, and professional and education fields. CoReasoningLab is in this sense not a computing-education tool but a generic instrument for any discipline in which a learner must specify an ill-defined task, judge a fallible solution, and steer it toward a better one.

***Table C1.*** *Disciplines covered by the released challenge content.*

| Area | Disciplines |
|---|---|
| STEM | Computer Science (algorithms); Physics (classical mechanics); Mathematics (linear algebra); Electrical Engineering (signal processing); Biology (molecular biology) |
| Social sciences | Economics (microeconomics); Political Science (international relations); Psychology (cognitive psychology) |
| Humanities and law | Philosophy (applied ethics); Law (constitutional law) |
| Professional and education | Business (organizational behavior); Education (educational technology) |

The four examples below are real ill-defined problems the platform generated, in applied ethics, physics, instructional design, and computing, each of which a learner must Frame before the AI produces a flawed solution to Judge and Steer:

- **Applied ethics.** "As a member of the AI Ethics Committee at a technology company, you have been tasked with developing a set of ethical guidelines for the deployment of a new AI system."
- **Physics.** "As part of a project for a local amusement park, design a new ride that involves a projectile-motion component, and analyze the outcomes."
- **Instructional design (education).** "Our organization is developing a new training program on effective communication skills. Design a backward lesson plan."
- **Computer science.** "Our company is building a navigation app. Design an algorithm that computes the shortest path between two locations on a road network with varying travel speeds, possible road closures, and concurrent route requests."

The deliberate gaps differ by discipline (a missing fairness criterion in the ethics task, an unstated launch or safety constraint in the physics task, undefined learning objectives in the lesson-plan task), but the learner's work is identical across all of them: specify the task (Framing), evaluate the AI's flawed solution (Judging), and redirect it toward an adequate one (Steering). Full per-discipline session logs ship with the platform.